\pdfoutput=1

\documentclass[11pt]{article}

\usepackage{EMNLP2022}

\usepackage{times}
\usepackage{latexsym}

\usepackage[T1]{fontenc}

\usepackage[utf8]{inputenc}

\usepackage{microtype}

\usepackage{inconsolata}

\usepackage{enumitem}
\setitemize{noitemsep,topsep=0pt,parsep=0pt,partopsep=0pt}
\definecolor{cite}{rgb}{0.6,0.6,1.0}

\definecolor{todo}{rgb}{1,0.5,0}

\usepackage{booktabs}
\usepackage{graphicx}
\usepackage{float}
\usepackage{tabularx}
\usepackage{soul}

\title{Yes-Yes-Yes: Proactive Data Collection\\ for ACL Rolling Review and Beyond}

\author{Nils Dycke$^*$, Ilia Kuznetsov$^*$, Iryna Gurevych\\ 
Ubiquitous Knowledge Processing Lab (UKP Lab)\\
Department of Computer Science and Hessian Center for AI (hessian.AI)\\
Technical University of Darmstadt \\
\texttt{ukp.informatik.tu-darmstadt.de}}

\newcolumntype{L}[1]{>{\raggedright\arraybackslash}m{#1}}
\newcolumntype{C}[1]{>{\centering\arraybackslash}m{#1}}
\newcolumntype{R}[1]{>{\raggedleft\arraybackslash}m{#1}}

\begin{document}

\maketitle
\begin{abstract}

The shift towards publicly available text sources has enabled language processing at unprecedented scale, yet leaves under-serviced the domains where public and openly licensed data is scarce. Proactively collecting text data for research is a viable strategy to address this scarcity, but lacks systematic methodology taking into account the many ethical, legal and confidentiality-related aspects of data collection. Our work presents a case study on proactive data collection in peer review -- a challenging and under-resourced NLP domain. We outline ethical and legal desiderata for proactive data collection and introduce \mbox{\emph{"Yes-Yes-Yes"}}, the first donation-based peer reviewing data collection workflow that meets these requirements. We report on the implementation of Yes-Yes-Yes at ACL Rolling Review\footnote{Code available at: \url{https://github.com/UKPLab/openreview-licensing-workflow}} and empirically study the implications of proactive data collection for the dataset size and the biases induced by the donation behavior on the peer reviewing platform.

\end{abstract}

\section{Introduction}

Empirical NLP is shaped by its data sources. %
While early work mostly considered \emph{canonical} sources like newswire \cite{wsj}, the last decade has been marked by the transition to fortuitous \cite{plank}, \emph{found data} openly available for collection. The rise of crowdsourcing platforms also made it possible to \emph{generate} massive amounts of textual data on demand \cite{snli, textgen}, feeding novel NLP developments in both task-specific problem engineering and general-purpose representation learning.

\begin{figure}[t!] 
	\centering
	\includegraphics[width=\linewidth]{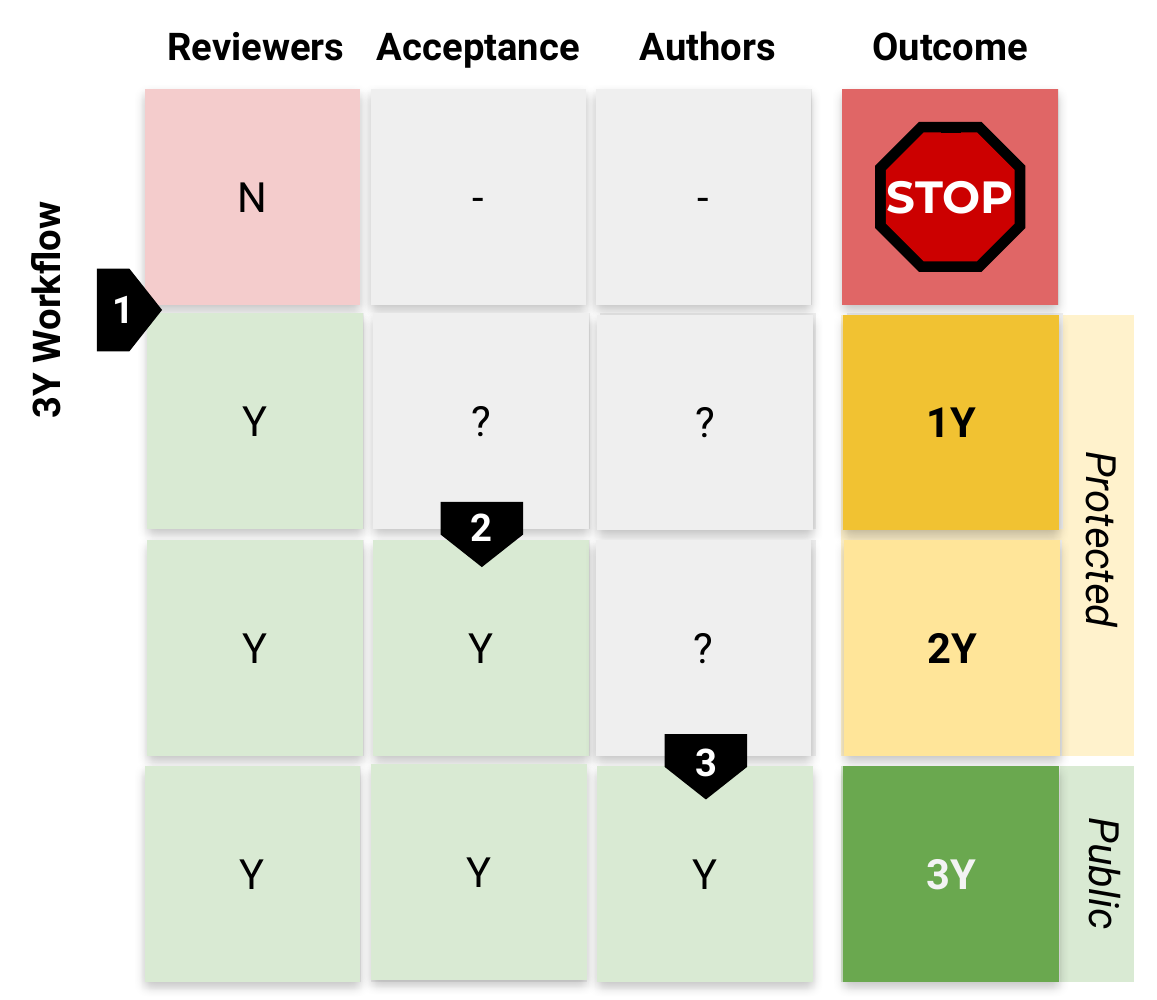}
	\caption{The Yes-Yes-Yes workflow: data collection is conditioned on participants' consent (1, 3) and on the confidentiality status of the publication (2). Only data that fulfills all conditions is added to public datasets.}%
\label{fig:3y-component}
\end{figure}

Yet problems persist with mainstream approaches to data collection. Canonical data is not representative. Not every language variety can be "found", and, by far, not every found text can be used for research \cite{rogers-data}. Generating texts "on demand" is prone to artifacts \cite{gururangan-etal-2018-annotation} and might be challenging in expert domains, like scientific or clinical text. As result, existing collection strategies in NLP leave under-serviced many domains and application scenarios that crucially depend on closed data or data not yet cleared for research use. 

Given the raising attention to the development of specialized NLP applications \cite{transnlp}, the ability to acquire data from the domains of interest on-demand becomes crucial. To that end, a promising but not systematically explored alternative to canonical, found, and generated source data is \emph{proactive}, targeted data collection, where texts are harvested from a previously closed process in the domain of interest, and made available for research. This requires careful consideration of the text production process in the target domain, as well as many ethical, legal, confidentiality-related and other aspects of data collection. In line with the recent work in responsible data handling for NLP \cite{rogers-data, bender-ds} and previous discussions on data collection in the digital mental health domain \cite{resnik}, our work explores practical implications of proactive data collection from a previously under-resourced domain: peer review.

Peer review is the cornerstone of academic quality control. The ever-increasing submission rates expose the weaknesses of this process%
, motivating the first generation of computational studies in peer review %
 within and beyond NLP \cite[etc.]{kang2018dataset, ampere, dycke2021ranking, reviewadvisor, herding, resub}. Such studies crucially depend on the availability of peer reviewing data. Yet this data is hard to come by and is associated with a range of ethical, confidentiality and copyright issues, and while current methodological advances in peer review processing show great promise, a solid data foundation for the study of peer review is still lacking. 
This makes peer review an excellent target for proactive data collection. In this work we:

\begin{itemize}
    \item outline the challenges and trade-offs associated with the peer reviewing data collection;
    \item propose \emph{Yes-Yes-Yes} (3Y) -- a generic data collection workflow to address those challenges;
    \item report on the instantiation of 3Y at ACL Rolling Review (ARR)\footnote{\url{https://aclrollingreview.org}} and examine the selectivity, bias and donation behavior in our workflow; %
    \item provide an open implementation of the proposed workflow for any research community that uses OpenReview\footnote{\url{https://openreview.net}}.
\end{itemize}
We highlight that this work is about data collection methodology and not about the dataset, which is subject to a subsequent study focused on peer review. Here, using peer review as an example domain, our goal is to spark the discussion on \emph{systematic approaches to proactive data collection from closed domains} in general, by outlining the challenges, proposing the workflow, and discussing the practical implications of ethical data collection. From the perspective of meta-science and scientific policy research, our work addresses the \emph{need for evidence-based empirical study of peer review} by proposing a workflow for ethical peer reviewing data collection which can be adapted to other communities. We elaborate on the underlying peer review system and publication culture at ARR in the appendix \ref{appendix} to contextualize our findings within the broader science landscape.

\section{Background}

\subsection{Text Sources in NLP}

Early work in NLP focused on few canonical text collections that were widely reused across studies. Yet, while core linguistic phenomena like POS are present in any domain, the NLP models of these phenomena suffer from domain shift, and as we move towards application-oriented NLP, target phenomena themselves become domain-dependent.
The ability to acquire text beyond canonical collections is thus critical for the success of both core and applied NLP. One strong alternative to canonical data is found data that emerges as a side-product of text communication and is readily accessible, e.g. Wikipedia, scientific publications, books, etc. %
Yet, not every text type can be found, and many specialized and rare discourse types are under-represented in NLP. Moreover, not every found text can be used for research, and an active line of work in NLP is concerned with the ethical, legal and privacy-related aspects of data collection and reuse \cite{rogers-data, bender-ds}. While it is possible to avoid some of these challenges by generating text on demand, this approach is prone to artifacts and limited both in scale and in the kinds of texts that can be created this way. %

Openly available texts only constitute a minor fraction of all texts produced. We claim that much of this data can be made available for research via proactive, donation-based data collection. 
Any restrictions imposed on the data naturally introduce bias. While recent work in NLP lays out the general principles for ethical data collection, the practical workflows that put it to use are missing, and it remains unclear how legal, ethical and other limitations shape the resulting data. Our work aims to bridge this gap by proposing a proactive data collection workflow and analysing its effects on data in the domain of peer reviews.

\subsection{Peer Review}

Scholarly peer review is a structured process that involves a range of stakeholders and produces many textual artifacts. A common reviewing campaign involves authors submitting their draft to a reviewing committee represented by editors. The editors distribute the drafts to reviewers who provide their evaluation in form of a report. In an optional revision stage the authors might communicate with reviewers and update the draft, and editors might produce meta-reviews to help decision-making. %
Based on the evaluation, the work is accepted or rejected; accepted work is subject to official publication. Peer review is often anonymized: the reviewer identities are hidden from the authors (single-blind), and the author identities might be hidden from the reviewers (double-blind).

Reviewing quality and efficiency are of paramount importance to maintain the integrity of science. Yet issues persist in both dimensions: reviewers are prone to a range of biases and strategic behaviors \cite{bias1, lee_bias_2013, strategicb, resub}, fall back on superficial heuristics \cite{rogers-pr}, and reviewing itself takes a lot of time and effort \cite{gspr}. This motivates computational study of peer review: \newcite{kang2018dataset} introduce PeerRead -- a corpus composed of openly available reports and drafts -- and report experiments on paper acceptance and aspect score prediction; \newcite{ampere} investigate argumentation in peer review reports; \newcite{ape} study the correspondences between review reports and author responses; \newcite{gao2019does} investigate the effect of rebuttal on evaluation; \newcite{dycke2021ranking} learn to rank papers based on review reports and scores; \citet{reviewadvisor} explore fully-automatic review report generation based on submission drafts.

\subsection{Status of Existing Peer Reviewing Data}

Computational research in peer review critically depends on the availability of open peer reviewing data. Existing studies in NLP for peer reviews build almost exclusively on two data sources: the reviews from the International Conference on Learning Representations (ICLR) available via the OpenReview platform, and reviews for accepted papers at the Conference on Neural Information Processing Systems (NeurIPS) 
available via the conference website\footnote{\url{https://iclr.cc}, \url{http://neurips.cc}}. Both ICLR and NeurIPS represent specialist communities in neural network and representation learning research -- a narrow sample given the widespread use of peer reviewing across scientific fields. While peer review in NLP and computational linguistics conferences has been previously studied \cite{kang2018dataset,gao2019does}, publicly available data is scarce.

Peer review is a challenging case for data collection. At the time of writing, neither ICLR nor NeurIPS provide information on the authors' and reviewers' consent for processing their peer reviewing data, nor specify the conditions of data processing by third parties. Yet peer reviewing data is personal and confidential, and hereby legally requires consent or other grounds for processing (see Section \ref{s:problems}). Publishing of, and attaching license and copyright to peer reviewing data is non-trivial and requires careful consideration of authorship and attribution. Yet none of the published datasets of peer reviews (incl. PeerRead \cite{kang2018dataset}, AMPERE \cite{ampere}, APE \cite{ape} and ASAPReview \cite{reviewadvisor}) attach clear license to the source or to the derivative annotated data, rendering the conditions of data re-use unspecified. All in all, the current ad-hoc approach to peer reviewing data collection in NLP shares a range of risks associated with the research use of found data in general, and the lack of standard data collection workflows results in a major overhead for individual data collection efforts. Our work aims to address these issues by proposing a general-purpose workflow for peer reviewing data collection built around a few key data collection principles that we outline next.

\section{Problem Dimensions} \label{s:problems}

We use \emph{peer reviewing data} as an umbrella term for the artifacts produced during peer review, and limit our discussion here to submission drafts and review reports.%
We distinguish between \emph{metadata} (numerical scores, track, paper format, etc.) and \emph{textual data}, and focus our discussion on the latter. Textual data falls under the EU General Data Protection Regulation (GDPR) definition of \emph{personal} data as "any information relating to an identified or identifiable natural person"\footnote{Although GDPR is EU-based, it is considered the best practice for privacy-related legislation globally; GDPR regulates processing of the EU subjects' data \textit{anywhere on Earth} and is thus almost certainly applicable to any major text collection; cf. an extended discussion in \cite{rogers-data}} -- the identities of the reviewers and authors are known to the editors, and remain potentially identifiable based on the writers' professional expertise and via author profiling. Although peer reviewers and authors rarely sign formal non-disclosure agreements, peer reviewing data is \emph{confidential}. Finally, most of the peer reviewing data is \emph{anonymous}, and only the editors know the identities of the participants.

\textbf{A. Collection strategy.} Personal data requires consent or other explicit grounds for processing. Two main approaches to obtaining consent are \emph{terms of service} (ToS) and \emph{donation}. ToS is a one-size-fits-all approach that requires the users to agree to the terms in order to use a platform. Establishing universal ToS is a challenge that involves balancing interests of many stakeholder groups, at the risk of losing participants who disagree with data collection. In a donation system, the decision to contribute data is made individually. Although technically more intricate, donation-based collection allows participants who do not wish to contribute to still use the platform. Donation-based approach might introduce participation bias into the data \cite{keeble2015choosing,slonim2013opting}.

\textbf{B. Stakeholder involvement.} Who has the authority to consent for what data? This question is not trivial: for example, in peer reviewing data, while a reviewer might agree to publishing their reports, the authors might object, not only due to potential negative reviews, but also due to the risk of leaking unpublished ideas and results pre-publication. Ideally one would want all involved stakeholders to consent; yet, increasing the number of involved parties can substantially reduce the amount of collected data and augment the bias.%

\textbf{C. Licensing.} Liberal data licensing allows the community to build upon prior data and ensures replicability. Creative commons (CC) is a popular licensing choice for NLP datasets that supports additional restrictions on data sharing, adaption and commercial use. Most CC licenses require attribution -- specifying the title, authorship and source of the data. Yet as most of the peer reviewing data is anonymous, it cannot be directly attributed to its authors, and declaring the work public domain (CC0) leaves the data reuse entirely unregulated, incl. commercialization and claiming copyright and attaching restrictive license to data derivatives.

\textbf{D. Anonymity.} During peer review, the identities of the authors are hidden to maintain the objectivity of review; anonymizing the reviewers aims to protect them from potential backlash. %
Yet, peer review is hard work; and if previously hidden review texts are to be made public as part of a dataset, the authors and reviewers should have an opportunity to be credited for their work, which effectively deanonymizes their contributions.

\textbf{E. Confidentiality.} Modern academia is highly competitive. %
Review reports %
often summarize papers in a way that enables a third party to appropriate the idea or to gain advantage due to the knowledge of unpublished results. Professional ethics prevent idea theft via peer review, as the identities of the reviewers are still known to the editors. Yet, if peer reviews are made available to the open public, this is no longer true, and access to unpublished research results 
presents a confidentiality risk.

\section{The Yes-Yes-Yes Workflow}
\subsection{Design} \label{s:design}
The aforementioned problem dimensions inform the design of our proposed workflow. We aim to grant key stakeholders -- {authors} and {reviewers} -- extensive control over their data, while maximizing the value of the resulting data to researchers. Both goals should be attained while ensuring least interference with the peer reviewing campaign and avoiding pressure on the {stakeholders}. We focus our data collection on {drafts} and review {reports}.

We collect data on a \textbf{donation basis} (Section \ref{s:problems}.A) and analyze the resulting participation bias. The \textbf{primary contributor} of the data is always the stakeholder producing the artifact (B.); this means, {drafts} must be donated by the {authors}, and the {reports} by {reviewers}. The stability and availability of the dataset is crucial for replicability of research results \cite{zubiaga2018longitudinal, rogers-data}, but simple consent does not guarantee data persistence, as \emph{it can be withdrawn}, which would in turn require modification of the underlying data; it is preferable to perform \textbf{license transfer} (C.): as long as the license conditions are met, the license cannot be revoked, and the research dataset remains stable. Reviewers and authors must have an opportunity to explicitly \textbf{request attribution} (D.), with the identity anonymous by default.
Finally, to account for the confidentiality (E.), permission to make {reports} public must be obtained from the {authors} (also, B.), and only material associated with \textbf{accepted publications} should be publicly released. Reports for which only {reviewers} opt-in can be subject to research but should not be made publicly accessible. %
\subsection{Workflow} \label{s:workflow}
Based on these design decisions, we define the Yes-Yes-Yes workflow (3Y-Workflow) for peer reviewing data collection at ARR (\autoref{fig:3y-component}) as a three-step decision process synchronized with the underlying peer reviewing campaign and applied on per-paper and per-reviewer basis. The workflow yields three possible \textbf{outcomes}: \emph{no data} collected (default), data added to a \emph{protected dataset} (potentially available for research, but not public), and data added to a \emph{public dataset}. In all cases, the resulting data is anonymous unless credit was explicitly requested by data contributors. In the following, we describe each step of the workflow in detail.

\textbf{1Y: Yes by the {Reviewers}.}
First, each {reviewer} decides on contributing their reports in the given reviewing campaign. To minimize the communication overhead, reviewers make a decision whether to donate all their reviewing reports, in bulk. To contribute, a reviewer signs a \textit{review report license agreement} with optional attribution (see \ref{as:rev_license}). This means that reviewer names are \emph{not} collected unless they explicitly request this. The donation can be made any time between submission and acceptance decisions. The reviewers are explicitly informed about the risks of future authorship attribution via profiling techniques (see \ref{as:rev_risk}). If the {reviewer} should not explicitly give the "first Yes", their reports are discarded from the data collection pipeline. Donated reviews become part of the protected dataset and the workflow proceeds.%

\textbf{2Y: Yes by the {Editors}.}
Next, we consider the acceptance decision on the submissions.
If editors accept a draft for publication ("second Yes") and its reviewers agreed to donate their peer review reports in the previous step, the workflow continues. The resulting 2Y reviews refer to published papers and thereby do not leak unpublished results. If the paper is not accepted, the reviews remain part of the 1Y protected dataset.

\textbf{3Y: Yes by the {Authors}.}
Finally, after the acceptance decisions are known, the {authors} of accepted papers are asked to contribute their drafts via a \textit{paper license agreement} (see \ref{as:aut_license}), as well as allow publication of the associated review reports from the {reviewers} that gave their "Yes" in the first step. By default, the authors opt-out all reviewing data associated with their draft (no draft and no reviews included). If they choose so, they can either donate just the paper draft or both the draft and its associated reviews. %
If the {authors} donate their data ("third Yes"), the draft and its donated review reports become eligible for the public dataset, 3Y. Otherwise, previously donated review reports for accepted papers remain in the protected 2Y dataset.

\textbf{Protected data} in the 1Y and 2Y datasets is confidential by design: it comprises anonymous review reports and metadata of agreeing reviewers (1Y) of both accepted and rejected papers (2Y). Due to confidentiality, it may be used to calculate statistics or to quantify biases, but cannot be made public. In our analysis below, we solely rely on non-sensitive numerical statistics from the protected data, and in the current implementation of the 3Y-Workflow at ARR, protected data is not collected, see Section \ref{sec:disc} for discussion.

\section{Implementation at ARR}

\subsection{ACL Rolling Review}
ACL Rolling Review (ARR) is an initiative in the ACL community that decouples peer review from publication and replaces the traditional, per-event reviewing campaigns with a single, journal-style reviewing process. ARR was launched in May 2021 and serves as the main reviewing platform for multiple major ACL conferences, including the Annual Meeting of the ACL\footnote{\url{https://www.2022.aclweb.org}}.
ARR operates in monthly cycles: during each cycle, the {authors} might submit their work to ARR; the draft is evaluated by {reviewers}; based on the evaluation, action editors decide whether the draft has passed peer review. If the evaluation is positive, the draft can be committed to one of the ACL conferences where program chairs (equivalent to {editors}) make the final decision to publish the work. If a draft is not accepted at a conference, it can be revised and resubmitted to ARR in next iterations. For the 3Y-Workflow, only the publication decision by the program chairs is relevant. %

ARR presents a unique opportunity for the study of peer review in the ACL community and beyond. The ever-increasing submission rates at ACL provide a steady source of reviewing data, and unified reviewing workflow, protocols and forms minimize the effects of a particular reviewing campaign configuration on the process. Consequentially, ARR is also highly suitable for the experimental study of proactive data collection in general, as we can vary collection configurations over time within a mostly fixed context of source data generation. In addition, the use of the open-source OpenReview platform makes it easy to automate many aspects of data collection, from sending out reminders to secure data filtering. With kind permission and support by the editors-in-chief and the technical team, we have implemented the 3Y-Workflow at ARR.

\subsection{Implementation Details} 

To minimize interference of the data collection with the peer reviewing campaign, our implementation relies on the built-in \texttt{Task} feature of OpenReview: optional data donation is seamlessly integrated as part of the reviewing process along with other tasks, like review submission. To enable future research on the collected data while preventing uncontrolled re-use and redistribution, we attach the Creative Commons \mbox{BY-NC-SA 4.0} License to the data which allows future users to share and adapt the data as long as it is attributed (BY), only used non-commercially (NC) and is shared under the same licensing conditions (share-alike, SA)\footnote{\url{https://creativecommons.org/licenses/by-nc-sa/4.0/}}. To avoid the pitfall of reviewing data being non-attributable due to anonymity, we ask the contributors to perform  \emph{license transfer} in which the copyright for the data is transferred to the ACL (similar to ACL Anthology\footnote{\url{https://aclanthology.org}} publications), while the data creators might still get attributed if they explicitly wish to reveal their identity.

To maintain the confidentiality of unpublished work, we currently opt to make public exclusively the peer reviewing data for papers that are later accepted at a venue and officially published. Our implementation of the 3Y-Workflow is openly available, making the data extraction code base transparent and allowing to easily set up the workflow for any new OpenReview-based reviewing campaign independent of the venue or research field. %

\section{Analysis}
\label{sec:analysis}

With the implementation of the proposed workflow at ARR, we can study the effects of donation-based proactive data collection on the resulting dataset composition, as well as the donation behavior across the users of the platform. For our analysis, we focus on the publications that were accepted at the 60th Annual Meeting of the ACL (ACL-2022) -- the first major conference to employ ARR as its main reviewing platform. As over $98\%$ of the papers accepted at ACL-2022 were submitted to September, October and November 2021 cycles of ARR, we consider those as the full dataset (\textbf{ARR-all}). We then focus the analysis on the metadata of the donated peer review reports from the subsets \textbf{ARR-1Y} (reviewers agree), \textbf{-2Y} (paper accepted) and \textbf{-3Y} (authors agree). For context, we bring in two prior datasets in the NLP domain. The \textbf{ACL-2018} \cite{gao2019does} was collected during the 56th Annual Meeting of the ACL; consent collection during the peer reveiwing process and the lack of license transfer prevented the public release of the full dataset.
The \textbf{ACL-2017} portion of the PeerRead corpus \cite{kang2018dataset} includes reviews of submissions to the 55th Annual Meeting of ACL, for which both authors and reviewers agreed to share the review reports.

\begin{table*}[t]
	\centering
	\begin{tabular}{R{2.2cm}C{1.6cm}C{1.7cm}C{1.7cm}C{1.8cm}C{1.8cm}C{2.1cm}}
	    \toprule
	    & \multicolumn{4}{C{7.8cm}}{\textbf{Our Collection}} & \multicolumn{2}{C{4.2cm}}{\textbf{Previous Work}} \\
	    \cmidrule(lr){2-5} \cmidrule(lr){6-7}
	    
		& \textbf{ARR-all} & \textbf{ARR-1Y} & \textbf{ARR-2Y} & \textbf{ARR-3Y} & \textbf{ACL-2018} & \textbf{ACL-2017}$_\textbf{\tiny{PR}}$ \\
		\midrule
		\# Submissions & $3591$ & $2884$ & $923$ & $235$ & $1528$ & $137$\\
		\# Reviews & $11621$ & $5656$ & $1815$ & $463$ & $3875$ & $275$\\
		\# Reviewers & $4421\downarrow$ & $1916$ & $1073$ & $388$ & $1213$ & -\\
		\# Reviews per submission & $3.24^*$ & $1.96 \pm 0.85$ & $1.97 \pm 0.87$ & $1.97 \pm 0.90$& $2.52 \pm 0.67$ & -\\ 
		\# Reviews per reviewer & $2.63\uparrow$ & $2.95 \pm 1.66$ & $1.69 \pm 0.86$ & $1.19 \pm 0.43$ & $3.04 \pm 1.35$ & -\\
		\bottomrule
	\end{tabular} 
	\caption{Statistics of donated (ARR-\{1, 2, 3\}Y) and all (ARR-all) reviews for September, October, November 2021. Reviewer statistics for ACL-2018 from \newcite{dycke2021ranking}; statistics on ACL-2017 portion of the PeerRead corpus from \newcite{kang2018dataset}. $\downarrow$ upper bound, $\uparrow$ lower bound, \mbox{$\ast$ estimated} from total counts.}
	\label{t:datastats}
\end{table*}

\subsection{Selectivity of Decisions}

Donation-based data collection inevitably is selective. We compare the collected data to limited statistics derived from the complete data (ARR-all). Aggregate metadata is generally not considered private and – despite the lack of explicit consent – it is widely used for conference reporting\footnote{For ARR reporting see: \url{https://stats.aclrollingreview.org/}}. Still, we received explicit one-time permission from the ARR editors-in-chief to obtain numerical statistics from ARR-all. These statistics were computed without access to identifiers of submissions, reviews or reviewers, posing no privacy risk during their creation.

As \autoref{t:datastats} shows, ARR-all already exceeds ACL-2018 by size, and although not all of this data is donated, the public data (ARR-3Y) already includes more reviews than the ACL-2017 subset of PeerRead. ARR-1Y covers approximately half of the reviews in ARR-all; each further decision in the workflow reduces the number of reviews (i.e. in ARR-2Y and ARR-1Y) by a third, rendering the process \textbf{highly selective}.
This confirms the anticipated limitation of proactive data collection based on multi-stakeholder decision making.
Yet, as the workflow is \textbf{continuously applied} at further conferences and ARR cycles, the ARR-3Y subset of the data is likely to outgrow ACL-2018 and other existing datasets in course of time.

Dataset bias towards individual data creators has recently gained attention in literature  \cite{Bandy2021AddressingD}. Even without explicit authorship information, aggregate statistics allow us to study diversity in our peer reviewing data.
The total number of unique reviewers across cycles is not available for ARR-all, as each cycle is managed as a separate reviewing campaign: in other words, if an individual reviewed at ARR in September and October, they would be counted twice. Hence, \autoref{t:datastats} reports an upper bound ($\downarrow$) of the number of reviewers and a lower bound for the number of reviews per reviewer ($\uparrow$) for ARR-all. As our results in \autoref{t:datastats} show, the number of reviews per submission remains nearly constant throughout the decision steps, as reviews are sub-selected on a per-submission basis. However, the number of reviews per reviewer drops notably from ARR-1Y to ARR-3Y; while this limits analyses on a per-reviewer basis, this also shows that reviews in ARR-3Y originate from many different creators.

\subsection{Polarity Bias} \label{subsec:bias_scores}

\begin{figure}
    \centering
    \includegraphics[width=0.98\linewidth]{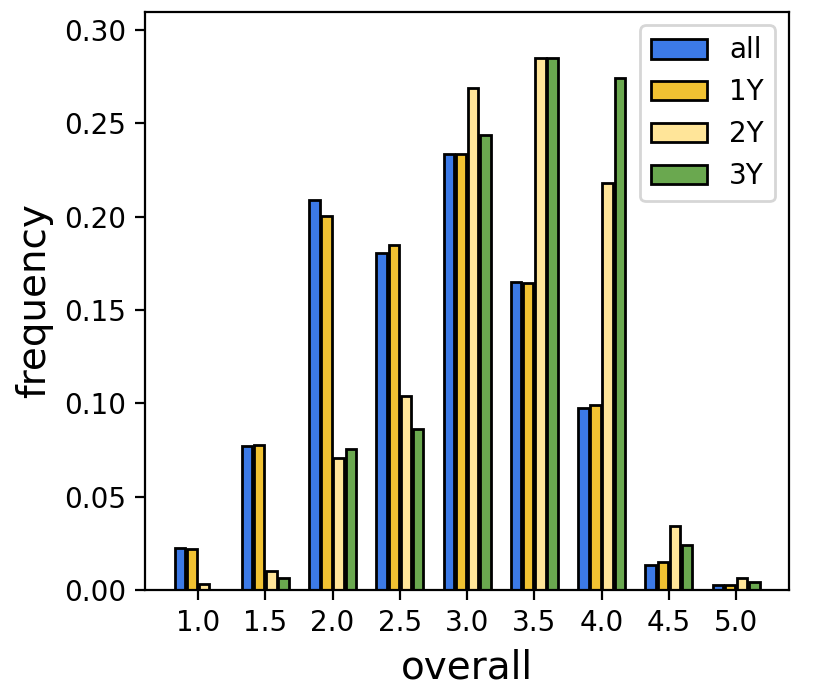}
    \caption{Overall score distribution in ARR-\{1, 2, 3\}Y and ARR-all reviews.}
    \label{fig:overall_score_dist}
\end{figure}

The decision to donate data is likely to correlate with a range of external factors. In peer review, one such potential factor is the review polarity -- both from the reviewer and from the author side. %
\autoref{fig:overall_score_dist} compares the distribution of overall scores in reviews of ARR-\{1, 2, 3\}Y and ARR-all. As it shows, the donated reviews cover a wide range of ratings, with the prevalent overall score around $3$ ("good"). The distribution of overall scores in ARR-1Y is nearly identical to ARR-all and resembles other computer science conferences \cite{ragone2013peer} and ACL-2018 \cite{gao2019does}. Thus, the reviewers' decision to donate data does not introduce substantial polarity bias. Yet, acceptance decision correlates with high scores, and for the roughly $25\%$ of submissions that are accepted (2Y) we observe a polarity bias. The final donation decision by authors of accepted papers (3Y) introduces no substantial further bias towards positive reviews; the review scores in ARR-3Y are only marginally more skewed towards positive scores than ARR-2Y with the most frequent score still at $3.5$ ("good"/"strong"). Thereby, we can conclude that the \textbf{conditioning on paper acceptance is the major source of polarity bias} in 3Y-Workflow.

Considering the interconnected nature of peer reviewing data, both reviewers and authors might base their donation decisions on all review scores of a submission rather than isolated reviews. To account for this, we study the bias towards papers with homogeneous ratings.
The agreement on overall scores by Krippendorff's $\alpha$ with ordinal metric \cite{krippendorff_content_1980} lies at $0.24$ for ARR-1Y, considerably lower than $0.34$ for ACL-2018 \cite{dycke2021ranking}; reviews for submissions with controversial ratings do seem to be donated, even when taking into account the different score scales of ACL-2018 and ARR. To investigate further, we consider the average standard deviation of overall scores per paper; if the overall scores per paper are homogeneous, this value is closer to zero. We observe only small differences between ARR-1Y ($0.259$), ARR-2Y ($0.253$) and ARR-3Y ($0.257$), showing that the \textbf{distribution of negative and positive reviews to papers is similar across subsets}.

\subsection{Donation Behavior} \label{subsec:participation}

A donation-based data collection workflow crucially depends on the {stakeholders}' participation, and we conclude our analysis with a brief overview of our observations related to donation behavior. Over the course of the considered three months, $2147$ responses to the donation request were collected from $4138$ active reviewers (each cycle of ARR treated individually). Among these responses $6.33\%$ explicitly disagreed to data collection, while the rest agreed. While the \textbf{majority of the reviewers prefer to stay anonymous}, a strong minority of $37.49\%$ \textbf{requested attribution}, showing the demand for getting credit for peer reviewing work.

By the implementation of the 3Y-Workflow at ARR, reviewers are free to sign the agreement before, during or after writing their review reports. Interestingly, $43.9\%$ of donating reviewers contributed their data before submitting their first review report of a cycle, while $40.47\%$ agree after their last reviewing report. This justifies leaving the decision timing up to reviewers and suggests that \textbf{the decision for donation is only weakly influenced by the outcome of the review}.%

Turning to the authors' participation, $29.53\%$ of the $999$ accepted paper drafts were donated, of which $87.79\%$ of the authors also agreed to the publication of the associated peer reviews. On the other hand, $37.34\%$ of authors explicitly disagreed to the collection. We note that despite the paper acceptance, author participation is nearly two times lower than reviewers': while private feedback from some of the authors revealed concerns over unfair negative reviews being published, the overall response rate of roughly $67\%$ suggests that better community engagement into collection would lead to higher participation and contribution rates.  %

\section{Discussion} \label{sec:disc}

3Y-Workflow is an example of proactive data collection; unlike found data, it allows the collector to interact with the text authors and clears the data for research use. The workflow lends itself to automation: apart from a few manual operations due to the current technical limitations of ARR, data collection can run continuously with minimal supervision, resulting in a steady supply of peer reviewing data from the NLP community, with the first data release currently in preparation. The open implementation allows adapting the workflow to other research communities at OpenReview.

Based on our experience, we determine four main directions for the future studies in proactive data collection. Developing better approaches to \textbf{stimulate participation} is crucial for fast dataset growth: this includes ensuring high visibility of the data collection effort, high transparency of the process, and deeper integration of data collection into reviewing process without interfering with the process itself. Our pilot investigation of \textbf{bias} in donation-based review collection revealed that some steps of the workflow indeed introduce polarity bias -- yet it remains unclear whether it affects NLP applications and what other biases are present. The study of bias requires access to the full reviewing data -- while we used anonymous numerical data from 1Y and 2Y as a proxy, deeper analysis would require access to review texts. While the 3Y-Workflow allows collection of protected data in theory (e.g. consented-to peer review texts blocked by the paper authors), \textbf{ensuring fair access to protected data} presents a technical, administrative and legal challenge: who provides and maintains the infrastructure for safe storage of the protected data? what can be done with this data? who can access the data and how is the access regulated? what are the consequences of protected data misuse and who would enforce them? A further community discussion is necessary to address those and other challenges and make protected data available for research.

Finally, from the legal perspective the replicability of research based on 3Y data is ensured via license transfer to a data controller -- in our case, the ACL. This approach has many advantages over publishing unlicensed data, but requires an external license holder and does not have a formal mechanism to ensure data withdrawal from derivative datasets without compromising replicability. The search for alternative \textbf{legal and technical frameworks} for publishing peer reviewing data constitutes another promising avenue for future studies.

Double-blind pre-publication peer review, as done at ARR, is an especially hard case for proactive data collection due to the confidentiality and anonymity. Not all proposed measures will be relevant for any proactive data collection campaign, even in peer review -- for example, other venues hosted by OpenReview make both accepted and rejected publications available, invalidating the confidentiality concern and rendering the 2Y step of our workflow redundant. Yet many of our findings in peer reviewing data collection at ARR point at the potential gaps in the proactive data collection methodology in general. Unlike canonical, found and generated data, proactive data collection demands systematic study of contribution behavior by the content creators. As most of the data to be collected proactively is user-generated personal data, solid ethical and legal frameworks are required to support its research use. As some of the data can never be made public, technical solutions for secure access to protected data are urgently needed to enable the study of bias in the new domains.

\section{Conclusion}

Given that most NLP tasks are either prone to domain shift, or are specific to a particular domain, or both, the future of NLP crucially depends on the field's ability to acquire new sources of textual data. We have presented "Yes-yes-yes", the first ethically sound, consent-driven data collection process for peer reviewing data and reported on its implementation at the ACL Rolling Review. We have further analysed the effects of ethical data collection strategies on dataset composition and found that different steps of our proposed data collection workflow indeed have quantifiable, systematic effects on the data. Yet, many questions remain open, from strategies for better community engagement, to the technicalities related to replicability and archival of the data. We hope that our study sparks the systematic discussion on proactive data collection strategies as a viable alternative to canonical and found data in NLP, in the domain of peer reviewing, and beyond.

\section*{Limitations}

In this work we have addressed proactive data collection in the peer reviewing domain. While our focus on a particular domain, research community and reviewing workflow made our study feasible in the first place, it inevitably limits the generality of our findings. A systematic study of donation behavior and data collection workflows would allow to evaluate our findings in the general case. Such systematic study requires us to determine parameters that can influence the outcome of data collection, and we outline some of them below.

In the peer reviewing domain, our study is limited in several important ways. First, we consider a specific, double-blind pre-publication workflow -- yet alternative workflows exist, incl. single-blind review, open review and post-publication review, and we expect the \emph{parameters of the peer reviewing process} to interfere with the willingness to donate the data, with the stance towards anonymity, and with requirements towards confidentiality. Second, we investigate a \emph{particular research community} in natural language processing -- yet peer review is used in most fields of science that differ in terms of their reviewing and community culture. Liberal \emph{licensing of the final publications} at ACL facilitates data collection as submission drafts published under an open license do not conflict with the final publication conditions. Yet, fields with more restrictive publication an dissemination standards would need to find a compromise between producing open research data and restrictive, potentially paid distribution of the final publications. ACL Rolling Review is a \emph{continuous reviewing process} -- meaning that the data size, and thereby participation rate -- are of secondary importance, as the necessary amount of data can be collected over time. Yet, when tackling one-time reviewing campaigns, additional work needs to be invested to maximize participation.

In context of proactive data collection in general, our findings would need further validation in other application scenarios and communities. We now briefly review the main ideas and findings of our work and discuss the extent to which they apply generally. We believe the proposed \emph{problem dimensions} (Section \ref{s:problems}) to apply to a wide range of scenarios outside peer reviewing domain. We deem our core \emph{design principles} (Section \ref{s:design}) applicable to a wide range of settings in peer review and beyond -- yet it is imaginable that additional constraints emerge in new domains, e.g. when working with sensitive personal data. Our specific \emph{data collection workflow} (Section \ref{s:workflow}) is tailored towards peer reviewing at ARR -- we expect it to be applicable to most peer reviewing campaigns with small modifications, yet it might need substantial adaptation for application outside peer review, as the text production process and the requirements change. With respect to our main findings (Section \ref{sec:analysis}), we expect \emph{high selectivity} to hold in most non-open peer reviewing environments -- yet our experimental setup does not allow us to distinguish between the explicit refusal to contribute and the gaps in outreach. Regarding \emph{bias}, we find that the reviewers' decision to donate the data does not substantially affect the score distribution -- which can be attributed to reviewers donating (or not donating) all their reviews in bulk, the strategy that we employed to simplify the data collection setup. We note that metadata that allows to study bias is a limiting assumption and might not be available in other data collection scenarios. We believe the score bias introduced by the filtering based on paper acceptance to be persistent in all datasets of peer reviews. The importance of this bias to NLP applications remains an open question. We believe \emph{demographics and the culture} of a particular user community, as well as \emph{timing and complexity} of the contribution process to be major factors in data donation, and see studies in data donation behavior as a viable future research direction.

\section*{Ethics Statement}

Our work directly contributes to the discussion of responsible data handling in NLP. While this paper does not introduce a dataset, the proposed workflow is designed to generate data that adheres to the current state of the art in handling personal data, liberal data licensing, as well as addresses the issues related to data confidentiality and anonymity. We believe that our discussion can be further refined, especially with respect to post-publication data withdrawal and protected data handling, and leave this for future studies. We believe the resulting data to be representative of its source population, namely the ACL community, to the degree that particular demographic characteristics systematically contribute to the willingness to donate the data. No annotators have been employed in production of the dataset and no demographic characteristics are collected or used in the study. For protected portions of the data, we only extract non-textual score statistics and do not utilize draft or review texts in any of the experiments. While this work does not come with a new dataset, we take additional consideration in treating anonymity and confidentiality of the contributed texts to minimize the possible harm from the future dataset use. Given the availability of other datasets of peer reviews, we do not believe that our data source introduces new, potentially harmful applications of NLP in the peer reviewing domain. Instead, it promotes fair, consent-based data collection that should enable reproducible and ethically sound NLP for peer review processing in the future. With the field shifting towards responsible handling of data, we deem it crucial to realistically highlight the practical implications of ethical data handling on the underlying datasets to guide future research in ethical data collection in NLP.

\section*{Acknowledgements}
We express our sincere gratitude to all parties providing support and advice during the realization of the peer review data collection at ACL Rolling Review. The data collection would not have been possible without the discussion and approval by the ACL Committee on Reviewing in 2021 chaired by Hinrich Schütze, and we are grateful to everyone who reached out to us during the pilot stages of the project to make suggestions and express their concerns -- some of which we could address, others left for the future iterations of the 3Y-Workflow. We thank the editors-in-chief and the technical team of ACL Rolling Review for their support during this ongoing data collection effort; with a special thanks to Amanda Stent and Sebastian Riedel. Finally, we thank Dorothy Deng for legal counseling and the specification of the review report and paper draft license agreement texts.\\
This research work is part of the InterText initiative\footnote{\url{https://intertext.ukp-lab.de}} at the UKP Lab. It has been funded by the German Federal Ministry of Education and Research and the Hessian Ministry of Higher Education, Research, Science and the Arts within their joint support of the National Research Center for Applied Cybersecurity ATHENE.

\bibliographystyle{acl_natbib}
\bibliography{emnlp2022}
\cleardoublepage
\appendix
\section{Appendix} \label{appendix}

\subsection{ACL Publication Norms} \label{as:pubnorms}
To contextualize our findings and enable the application of our workflow in other research fields, here we report key information on the publication norms in the Association for Computational Linguistics (ACL) community and its conferences, in aggregate referred to as *ACL.

\paragraph{ACL Conferences}
ACL is a major, world-wide professional organisation in the field of Computational Linguistics and Natural Language Processing.
Following a general trend in machine learning and NLP, the ACL community uses fast-paced conference-based publishing, with some conference publications attaining similar level of prestige, visibility and impact as journal articles. ACL holds regular meetings worldwide, attracting thousands of submissions in cutting-edge NLP. *ACL conferences include but are not limited to the main ACL conference, regional chapters (AACL, EACL, NAACL) and the conference on Empirical Methods in Natural Language Processing (EMNLP). Since 1979 the main ACL conference have been held annually\footnote{For an overivew see: \url{https://aclanthology.org/}}.
\paragraph{Peer Review at *ACL}
*ACL conferences are competitive, and the explosive growth in submission rates puts a strain on the peer reviewing processes at *ACL. As a dynamic, multi-disciplinary, fast-growing field, NLP faces many challenges related to peer reviewing, incl. bias, mixed review quality, and high effort associated with reviewing. Systematic, evidence-based study of peer review requires data. While *ACL conferences produce large amounts of peer reviewing data, peer review at *ACL is double-blind, and in the past peer reviews have not been published systematically.

\paragraph{ACL Rolling Review} Until ACL Rolling Review (ARR) has been released as a pilot project in 2021, each *ACL conference employed an individual peer reviewing campaign. ARR introduced a unified journal-style reviewing workflow aiming to reduce reviewing overhead in the community: submissions are reviewed and then accepted, rejected or resubmitted to the next ARR iteration. Once a paper passes peer review, it can be committed to be published and presented at a conference of choice; the conference program chairs make the final decision whether the manuscript is published at the selected venue. The editorial peer reviewing process for the monthly iterations at ARR is a four-step procedure:
\begin{enumerate}[noitemsep]
    \item The editors-in-chief desk reject papers violating policies on formatting or anonymization.
    \item Each paper is assigned to an action editor, who manages the reviewers for each of their assigned papers and makes the final recommendation.
    \item Each paper is assigned three to four reviewers based on a matching score from their researcher profile, but without bidding\footnote{Please refer to the repository of ARR for the details of the matching algorithm: \url{https://github.com/acl-org/arr-openreview}}.
    \item After the reviewers submitted their reports, the action editors make a decision for or against the revision of the paper.
\end{enumerate}

\paragraph{Norms of Paper Writing at *ACL}
Paper submissions adhere to a standardized \LaTeX-template and typically consist of eight pages for long papers and four pages for short papers. This ensures consistent formatting of the paper drafts. The language of the drafts is exclusively English, yet some conferences in the past encouraged additional submission of abstracts in other languages. Within the NLP community various paper types are distinguished, which are subject to different styles of reviewing\footnote{\url{https://coling2018.org/paper-types/}.}; some examples include resource papers, position papers or method papers.\\

\paragraph{Norms of Review Writing at *ACL}
ARR offers standartized peer reviewing forms. Reviews consist of a textual review report sub-divided in standardized fields and a range of review scores to ease their aggregation (for more details please refer to the exact form specified in Section \ref{as:review_form}). The language of reviews is exclusively English. Generally, peer reviewing at ARR is not impact-neutral, instead the relevance to *ACL conferences and the estimated audience are taken into account while writing a review. To contextualize reviewing practices, please refer to the following resources on best-practices in reviewing at *ACL:
\begin{itemize}
    \item \url{https://aclrollingreview.org/reviewertutorial}
    \item \url{https://acl2017.wordpress.com/2017/02/23/last-minute-reviewing-advice/}
    \item \url{https://2021.aclweb.org/blog/reviewing-advice/}
\end{itemize}

\newpage

\subsection{Review Report License Agreement} \label{as:rev_license}
To add more detail on the license transfer, we report the license agreement for reviewers as used in the 3Y-Workflow at ARR (status May 2022). We \ul{underline} passages that benefit from a more informal explanation and discussion provided in a footnote for each passage. We add this commentary based on the interaction with ACL legal counseling to make the given text more easily accessible to a broad audience without legal background; however, we do not claim legal interpretative authority of these passages. Likewise, there are no legal warranties with respect to the re-use of this license agreement in other collection efforts.

Along with this license agreement, reviewers were presented a disclaimer informing them about the purpose of the license transfer and potential risks of author profiling to infer their identity based on text.

\paragraph{Association for Computational Linguistics Peer Reviewer Content License Agreement}
Name of ACL Conference: \texttt{cycle name}\\
Peer Reviewer’s Name: \texttt{reviewer identity}\\
$^*$ Unless the peer reviewer elects to be attributed according to Section 2, the peer reviewer’s name
will not be identified in connection with publication of the Peer Review Content. If you wish to be
attributed, please check this box \fbox{$\phantom{3}$}.\\ \\
This Peer Reviewer Content License Agreement (“Agreement”) is entered into between the
Association for Computational Linguistics (“ACL”) and the Peer Reviewer listed above in
connection with content developed and contributed by Peer Reviewer during the peer review
process (referred as “Peer Review Content”).\\ \\
In exchange of adequate consideration, ACL and the Peer Reviewer agree as follows:
\begin{enumerate}
    \item Grant of License. Peer Reviewer grants ACL a worldwide, \ul{irrevocable}\footnote{Irrevocable license transfer is necessary to ensure dataset stability and sublicensing for derivative datasets. License transfer is also a common practice in the scientific publication landscape before including papers in conference proceedings our journals. Still, we design the collection and data storage to enable the removal of individual data points on request.}, and royalty-free
license to use the Peer Review Content developed and prepared by Peer Reviewer in
connection with the peer review process for the ACL Conference listed above, \ul{including but
not limited to text, review form scores and metadata, charts, graphics, spreadsheets, and any
other materials}\footnote{The broad types of data covered by this statement are designed to make the license agreement independent of specific review forms and hereby avoid constant changes induced by modifications to the underlying peer review system. As outlined below, this specifically excludes (meta-)data that would de-anonymize a reviewer by any means.} according to the following terms:
    \begin{enumerate}
        \item For Peer Review Content associated with papers accepted for publication, and subject
to the Authors permission, ACL may reproduce, publish, distribute, prepare
derivative work, and otherwise make use of the Peer Review Content,
and to sub-license the Peer Review Content to the public according to terms of the
\ul{Creative Commons Attribution-NonCommercial-ShareAlike 4.0 International License}\footnote{Until January 2022 the standard license for ACL publications was used, which permits commercial use. We opted to prohibit the commercial use of review data entirely and enforce this stricter license retroactively for all previous cycles.}.
\item For Peer Review Content associated with papers not accepted for publication, \ul{ACL
may use the Peer Review Content for internal research}\footnote{This statement refers to the protected dataset. The ACL as an organization may grant actors rights in its name; these actors are not explicitly specified in this agreement. While "internal research" prohibits the publication of the protected data, this term is also underspecified. Hence, we opt to consider the protected dataset exclusively for bias analysis in a secure environment and abstain from storing this data or conducting experiments beyond summarizing statistics on its meta-data}, program analysis, and record-keeping purposes.
Notwithstanding the foregoing, the Parties acknowledge and agree that \ul{this Agreement does
not transfer to ACL the ownership of any proprietary rights}\footnote{It is important to note that ACL does \textit{not} become the owner of the peer review data. It only receives the right to sub-license it under the given terms and conditions of this agreement. Ownership remains with the reviewers.} pertaining to the Peer Review
Content, and that Peer Review retains respective ownership in and to the Peer Review Content.
    \end{enumerate}
    \item Attribution and Public Access License.
    \begin{enumerate}
        \item The Parties agree that for purpose of administering the public access license, ACL will be
identified as the licensor of the Content with the following copyright notice:
Copyright © 2022 administered by the Association for Computational Linguistics (ACL)
on behalf of ACL content contributors: ... (list names of peer reviewers
who wish to be attributed), and other contributors who wish to remain anonymous.
Content displayed on this webpage is made available under a Creative Commons Attribution-NonCommercial-ShareAlike 4.0 International License.
    \item In the event Peer Reviewer intends to modify the attribution displayed in connection with
the copyright notice above, ACL will use \ul{reasonable efforts to modify}\footnote{As outlined earlier, we take measures to remove individual data points upon request by reviewers. The deletion of datapoints is, however, non-trivial: there is no leverage to remove datapoints from derivative datasets created by other parties after the initial publication of the dataset. Tracing back individual datapoints in derivative datasets is a complex problem that requires further investigation.} the copyright notice
after receipt of Peer Reviewer’s written request. Notwithstanding the foregoing, Peer
Reviewer acknowledges and agrees that any modification in connection with attribution
will not be retroactively applied.
    \item The Parties understand and acknowledge that the Creative Commons Attribution-NonCommercial-ShareAlike 4.0 International License is \ul{irrevocable once granted unless the licensee breaches the license terms}\footnote{This means, if the ACL should violate these terms, it loses the right of distribution of the signee's peer review data.}.
    \end{enumerate}
    \item  \ul{Warranty}\footnote{This statement is very similar to warranty guarantees provided by authors of papers when making their work public via publisher. Only original work owned by the signee is permitted.}. Peer Reviewer represents and warrants that the Content is Peer Reviewer’s
original work and does not infringe on the proprietary rights of others. Peer Reviewer further
warrants that he or she has obtained all necessary permissions from any persons or
organizations whose materials are included in the Content, and that the Content includes
appropriate citations that give credit to the original sources.
    \item Legal Relationship. The Parties agree that this Agreement is not intended to create any joint
venture, partnership, or agency relationship of any kind; and both agree not to contract any
obligations in the name of the other.
\end{enumerate}
Signature: \texttt{signature}, 
Date: \texttt{date}\\
Name Typed: \texttt{name}
\subsection{Disclaimer for Reviewers} \label{as:rev_risk}
To ensure that reviewers are aware of the risks associated with the donation of their (anonymous) reviewing data, the following disclaimer is presented along with the review report license agreement:\\

\texttt{
Your participation is strictly voluntary. By transferring this license you grant ACL the right to distribute the text of your review. In particular, we may include your review text and scores in research datasets without revealing the OpenReview identifier that produced the review. Keep in mind that as with any text, your identity might be approximated using author profiling techniques. Only reviews for accepted papers will be eventually made publicly available. The authors of the papers will have to agree to the release of the textual review data associated with their papers.
}
\newpage
\subsection{Paper License Agreement} \label{as:aut_license}
Here, we report on the license agreement for authors as used in the implementation of the 3Y-Workflow at ARR (status May 2022). As in the previous subsection, we \ul{underline} passages that benefit from a more informal explanation and discussion provided in a footnote for each passage. We focus on parts that deviate from the reviewers' license agrement.

\paragraph{Association for Computational Linguistics Blind Submission License Agreement}
Name of ACL Conference: \texttt{cycle name}\\
Blind Submission Paper Title: \texttt{title}\\
List Authors’ Names: \texttt{author identifiers}\\

$^*$ Authors names will not be shared with the peer reviewers during the peer review process
This Blind Submission License Agreement (“Agreement”) is entered into between the
Association for Computational Linguistics (“ACL”) and the Authors listed in connection with
Authors’ blind submission paper listed above (referred as “Blind Submission Content”).
In exchange of adequate consideration, ACL and the Authors agree as follows:

\begin{enumerate}
    \item Grant of License. After the peer review process is concluded and upon acceptance of the
paper, Authors grant ACL a worldwide, irrevocable, and royalty-free license to use the \ul{blind
submission paper version}\footnote{This includes only the submitted PDF and no other information. Specifically, supplementary materials like code or datasets are excluded.} (referred as “Content”). The foregoing license grants ACL the right
to reproduce, publish, distribute, prepare derivative work, and otherwise make
use of the Content, and to sub-license the Content to the public according to terms of the
Creative Commons Attribution-NonCommercial-ShareAlike 4.0 International License.
Notwithstanding the foregoing, the Parties acknowledge and agree that this Agreement does
not transfer to ACL the ownership of any proprietary rights pertaining to the Content, and that
the Authors retain their respective ownership in and to the Content.
\item Permission to Publish Peer Reviewers Content. After the peer review process is concluded
and upon acceptance of the paper, Authors have the option to grant ACL permission to publish
peer reviewer's content associated with the Content, which may include text, review form
scores and metadata, charts, graphics, spreadsheets, and any other materials developed by peer
reviewers in connection with the peer review process.\\
\fbox{$\phantom{3}$} Authors grant permission for ACL to publish peer reviewers content\\
\fbox{$\phantom{3}$} Authors decline to grant permission for ACL to publish peer reviewers content
\item  Attribution and Public Access License.
\begin{enumerate}
    \item The Parties agree that for purpose of administering the public access license, ACL will be
identified as the licensor of the Content with the following copyright notice:
Copyright © 2022 administered by the Association for Computational Linguistics (ACL)
on behalf of the authors and content contributors. Content displayed on this webpage is
made available under a Creative Commons Attribution-NonCommercial-ShareAlike 4.0 International License.
\item The Parties understand and acknowledge that the Creative Commons Attribution-NonCommercial-ShareAlike 4.0 International License is irrevocable once granted unless the licensee breaches the license terms.
\end{enumerate}
\item Effective Date. The grant of license pursuant to Section 1 and permission to publish peer
reviewers content pursuant to Section 2 becomes effective in the event Authors’ blind
submission paper is accepted for publication by ACL. If the blind submission paper is not
accepted, the Content and associated peer reviewers content will remain confidential and kept
for internal record-keeping purpose only.
\item Warranty. Authors represent and warrant that the Content is Authors’ original work and does
not infringe on the proprietary rights of others. Authors further warrant that they have
obtained all necessary permissions from any persons or organizations whose materials are
included in the Content, and that the Content includes appropriate citations that give credit to
the original sources.
\item Legal Relationship. The Parties agree that this Agreement is not intended to create any joint
venture, partnership, or agency relationship of any kind; and both agree not to contract any
obligations in the name of the other.
\end{enumerate}
By selecting 'On behalf of all authors, I agree' below, I confirm that all Authors have agreed to the above terms and that I am authorized to execute this Agreement on their behalf. Optionally, if you wish to transfer the license to the peer reviewing and blind submission data of all previous versions of this paper submitted to ARR, please select 'On behalf of all authors, I agree for all previous versions of this submission'.\\
\fbox{$\phantom{3}$} On behalf of all authors, I agree \\
\fbox{$\phantom{3}$} On behalf of all authors, I do not agree \\
\fbox{$\phantom{3}$} On behalf of all authors, I agree for this and all previous versions of this submission \\

Signature \texttt{signature}, 
Date \texttt{date} \\

Name (please print) \texttt{author's name}
\cleardoublepage
\subsection{Review Form at ARR} \label{as:review_form}
As a reference, we report on the review form used throughout the considered cycles of September, October, November at ARR. The following fields with given descriptions were presented to the reviewers.

\paragraph{Paper Summary}
Describe what this paper is about. This should help action editors and area chairs to understand the topic of the work and highlight any possible misunderstandings. Maximum length 20000 characters.

\paragraph{Summary Of Strengths}
What are the major reasons to publish this paper at a selective *ACL venue? These could include novel and useful methodology, insightful empirical results or theoretical analysis, clear organization of related literature, or any other reason why interested readers of *ACL papers may find the paper useful. Maximum length 20000 characters.

\paragraph{Summary Of Weaknesses}
What are the concerns that you have about the paper that would cause you to favor prioritizing other high-quality papers that are also under consideration for publication? These could include concerns about correctness of the results or argumentation, limited perceived impact of the methods or findings (note that impact can be significant both in broad or in narrow sub-fields), lack of clarity in exposition, or any other reason why interested readers of *ACL papers may gain less from this paper than they would from other papers under consideration. Where possible, please number your concerns so authors may respond to them individually. Maximum length 20000 characters.

\paragraph{Comments, Suggestions And Typos}
If you have any comments to the authors about how they may improve their paper, other than addressing the concerns above, please list them here.
Maximum length 20000 characters.

\paragraph{Overall Assessment}
\begin{itemize}
\item 5 = Top-Notch: This paper has great merit, and easily warrants acceptance in a *ACL top-tier venue.
\item 4.5 
\item 4 = Strong: This paper is of significant interest (for broad or narrow sub-communities), and warrants acceptance in a top-tier *ACL venue if space allows.
\item 3.5 
\item 3 = Good: This paper is of interest to the *ACL audience and could be published, but might not be appropriate for a top-tier publication venue. It would likely be a strong paper in a suitable workshop.
\item 2.5 
\item 2 = Borderline: This paper has some merit, but also significant flaws. It does not warrant publication at top-tier venues, but might still be a good pick for workshops.
\item 1.5 
\item 1 = Poor: This paper has significant flaws, and I would argue against publishing it at any *ACL venue.
\end{itemize}

\paragraph{Confidence}
\begin{itemize}
\item 5 = Positive that my evaluation is correct. I read the paper very carefully and am familiar with related work.
\item 4 = Quite sure. I tried to check the important points carefully. It's unlikely, though conceivable, that I missed something that should affect my ratings.
\item 3 =  Pretty sure, but there's a chance I missed something. Although I have a good feel for this area in general, I did not carefully check the paper's details, e.g., the math or experimental design.
\item 2 =  Willing to defend my evaluation, but it is fairly likely that I missed some details, didn't understand some central points, or can't be sure about the novelty of the work.
\item 1 = Not my area, or paper is very hard to understand. My evaluation is just an educated guess.
\end{itemize}

\paragraph{Best Paper}
Could this be a best paper in a top-tier *ACL venue?
\begin{itemize}
\item Yes
\item Maybe
\item No
\end{itemize}

\paragraph{Best Paper Justification}
If the answer on best paper potential is Yes or Maybe, please justify your decision.

\paragraph{Replicability}
Will members of the ACL community be able to reproduce or verify the results in this paper?
\begin{itemize}
\item 5 = They could easily reproduce the results.
\item 4 = They could mostly reproduce the results, but there may be some variation because of sample variance or minor variations in their interpretation of the protocol or method.
\item 3 = They could reproduce the results with some difficulty. The settings of parameters are underspecified or subjectively determined, and/or the training/evaluation data are not widely available.
\item 2 = They would be hard pressed to reproduce the results: The contribution depends on data that are simply not available outside the author's institution or consortium and/or not enough details are provided.
\item 1 = They would not be able to reproduce the results here no matter how hard they tried.
\end{itemize}

\paragraph{Datasets}
If the authors state (in anonymous fashion) that datasets will be released, how valuable will they be to others?
\begin{itemize}
\item 5 = Enabling: The newly released datasets should affect other people's choice of research or development projects to undertake.
\item 4 = Useful: I would recommend the new datasets to other researchers or developers for their ongoing work.
\item 3 = Potentially useful: Someone might find the new datasets useful for their work.
\item 2 = Documentary: The new datasets will be useful to study or replicate the reported research, although for other purposes they may have limited interest or limited usability. (Still a positive rating)
\item 1 = No usable datasets submitted.
\end{itemize}

\paragraph{Software}
If the authors state (in anonymous fashion) that their software will be available, how valuable will it be to others?
\begin{itemize}
\item 5 = Enabling: The newly released software should affect other people's choice of research or development projects to undertake.
\item 4 = Useful: I would recommend the new software to other researchers or developers for their ongoing work.
\item 3 = Potentially useful: Someone might find the new software useful for their work.
\item 2 = Documentary: The new software will be useful to study or replicate the reported research, although for other purposes it may have limited interest or limited usability. (Still a positive rating)
\item 1 = No usable software released.
\end{itemize}

\paragraph{Author Identity Guess}
Do you know the author identity or have an educated guess?
\begin{itemize}
\item 5 = From a violation of the anonymity-window or other double-blind-submission rules, I know/can guess at least one author's name.
\item 4 = From an allowed pre-existing preprint or workshop paper, I know/can guess at least one author's name.
\item 3 = From the contents of the submission itself, I know/can guess at least one author's name.
\item 2 = From social media/a talk/other informal communication, I know/can guess at least one author's name.
\item 1 = I do not have even an educated guess about author identity.
\end{itemize}

\paragraph{Ethical Concerns}
Independent of your judgement of the quality of the work, please review the ACL code of ethics (https://www.aclweb.org/portal/content/acl-code-ethics) and list any ethical concerns related to this paper. Maximum length \underline{10000} characters.

\end{document}